\PassOptionsToPackage{colorlinks=true, allcolors=blue}{hyperref}
\documentclass[pdflatex,sn-mathphys-num]{sn-jnl}% Math and Physical Sciences Numbered Reference Style
\usepackage{graphicx}%
\usepackage{multirow}%
\usepackage{amsmath,amssymb,amsfonts}%
\usepackage{amsthm}%
\usepackage{mathrsfs}%
\usepackage[title]{appendix}%
\usepackage{xcolor}%
\usepackage{textcomp}%
\usepackage{manyfoot}%
\usepackage{booktabs}%
\usepackage{algorithm}%
\usepackage{algorithmicx}%
\usepackage{algpseudocode}%
\usepackage{listings}%
\usepackage[colorlinks=true, allcolors=blue]{hyperref}
\begin{document}

\title{Deep Learning-Based Segmentation of Peritoneal Cancer Index Regions from CT Imaging}

\author*[1]{\fnm{Pieter C.} \sur{Gort}}

\author[1,2]{\fnm{Lotte J.S.} \sur{Fleurkens-Ewals}}

\author[1,2]{\fnm{Lenah D.} \sur{Kampmeijer}}

\author[1,2]{\fnm{Anna F.} \sur{van Herwijnen}}

\author[2]{\fnm{Marion W.} \sur{Tops-Welten}}

\author[1]{\fnm{Cris H.B.} \sur{Claessens}}

\author[2]{\fnm{Joost} \sur{Nederend}}

\author[2]{\fnm{Ignace H.J.T.} \sur{De Hingh}}

\author[3]{\fnm{Max J.} \sur{Lahaye}}

\author[2]{\fnm{Misha D.P.} \sur{Luyer}}

\author[1]{\fnm{Fons} \spfx{van der} \sur{Sommen}}

\email{\{p.c.gort, c.h.b.claessens, fvdsommen\}@tue.nl}
\email{\{lotte.ewals, marion.tops, lenah.kampmeijer, anna.v.herwijnen, joost.nederend\}@catharinaziekenhuis.nl}

\affil*[1]{\orgdiv{Department of Electrical Engineering}, \orgname{Eindhoven University of Technology}, \orgaddress{\city{Eindhoven}, \country{The Netherlands}}}
\affil[2]{\orgname{Catharina Hospital Eindhoven}, \orgaddress{\city{Eindhoven}, \country{The Netherlands}}}
\affil[3]{\orgname{Antoni van Leeuwenhoek Institute}, \orgaddress{\city{Amsterdam}, \country{The Netherlands}}}

\abstract{\textbf{Purpose:} Peritoneal metastases~(PM) are staged using the surgically determined Peritoneal Cancer Index~(PCI), which requires invasive laparoscopic assessment. Although CT is routinely used for preoperative evaluation, imaging-based assessment of PM extent remains challenging and is often less structured than surgical PCI scoring. A recent consensus study defined radiological PCI (rPCI) regions for cross-sectional imaging. We present the first deep learning approach to automatically segment 13~rPCI regions on CT.\\
\textbf{Methods:} 62~contrast-enhanced CT scans were retrospectively collected across the full PCI range. Each scan was annotated into non-overlapping rPCI regions by one researcher, reviewed by a second, with disagreements resolved by a radiologist. Using five-fold cross-validation, we compared nnU-Net and Swin UNETR with Dice, 95th-percentile Hausdorff distance (HD95) and Average Surface Distance (ASD). We introduce an anatomically constrained pipeline that trains on merged super-regions and splits them during post-processing using TotalSegmentator landmarks at the hips and the ligament of Treitz.\\
\textbf{Results:} On the identical 62-scan cohort, the baseline nnU-Net reached an overall Dice of 0.81 and outperformed Swin UNETR (0.76). The proposed anatomically constrained pipeline improved the overall Dice to 0.84 and reduced boundary error (HD95 $13.7 \to 11.8$~mm; ASD $4.1 \to 3.4$~mm), with the largest gains in the small-bowel regions, approaching the interobserver Dice of 0.87.\\
\textbf{Conclusions:} Automated rPCI region segmentation on CT is feasible and approaches interobserver agreement. Encoding anatomical boundary constraints substantially improves segmentation quality in the most challenging regions. This provides a reproducible foundation for non-invasive, imaging-based PCI assessment. The main limitations are the single-center cohort and the small interobserver subset. Code is available at: \url{https://github.com/PieterGort/rpci-region-segmentation}}


\maketitle

\section{Introduction}\label{sec:intro}
Peritoneal Metastases~(PM) refer to the spread of malignant tumors within the abdominal cavity lining, also called the peritoneum. PM arise most commonly from gastric cancer, colorectal cancer, and in women also ovarian cancer~\cite{Rijken2023OnMetastases}. Until recently, the diagnosis of PM was associated with a poor prognosis and was primarily managed with palliative care. In recent decades, the development of treatments for patients with PM has improved the prognosis. Treatment options have expanded with cytoreductive surgery combined with hyperthermic intraperitoneal chemotherapy, pressurized-intraperitoneal aerosol chemotherapy and intraperitoneal chemotherapy~\cite{Noiret2022UpdateReview, Guchelaar2023IntraperitonealMalignancies}. To evaluate treatment eligibility and monitor therapeutic response, it is essential to accurately determine the extent of PM~\cite{Noiret2022UpdateReview}.

The extent of PM is currently evaluated using diagnostic laparoscopy~(DLS), which is considered the gold standard for diagnosing PM~\cite{Hentzen2019RoleChemotherapy}. To quantify the severity of PM, the Sugarbaker's Peritoneal Cancer Index~(PCI) is most commonly used~\cite{Jacquet1996}. This system divides the abdominal cavity into 13~regions (numbered~0~to~12), after which each region is assigned a score of 0~to~3 based on the size of the largest tumor present in the region. The sum of all regional scores is the final PCI that ranges within~[0, 39]. This surgically determined PCI~(sPCI) provides physicians with a good prognostic indicator and is widely used in practice today, as well as in clinical trials~\cite{Guchelaar2023IntraperitonealMalignancies, Noiret2022UpdateReview, vantSant2020}. In addition, it is used to determine whether a patient is eligible for a certain treatment~\cite{Dohan2018PreoperativeOrigin}. Along with the expansion of treatment options, there is a growing need for non-invasive techniques to assess the presence and extent of PM before, during, and after treatment.

To assess the spread of the disease, patients with PM typically undergo a computed tomography~(CT) scan to assess the primary tumor and the presence of potential metastases. The assessment of PM on imaging is inherently challenging and careful radiological interpretation is required. Subtle lesions are easily overlooked, and compared to surgical assessment, the disease extent is often underestimated~\cite{Esquivel2010AccuracyStudy}. Furthermore, the sPCI system is not directly applicable to imaging-based assessment of the PCI. To facilitate a structured assessment of PM on imaging, a recent Delphi study involving 88~international experts in radiology, surgery, and gynecology proposed region definitions for a radiological PCI~(rPCI)~\cite{Tops-Welten2025DefiningStudy}. The study used multiple rounds of questionnaires aimed at defining region boundaries for rPCI assessment. Tops-Welten~\emph{et~al.}~\cite{Tops-Welten2025DefiningStudy} reported high consensus ($\geq$75\% agreement) for regions~0--8 and major consensus (60--75\% agreement) for regions~9--12, which encompass the small bowel. The rPCI regions provide anatomical landmarks required for reproducible imaging-based assessments of PM. Importantly, the rPCI regions are a delineation of the peritoneal cavity rather than organ-level segmentations: each region is defined by anatomical landmarks and surrounding organ surfaces, representing a spatial subdivision rather than a segmentation of the organs themselves.

In this work, our objective is to take advantage of the aforementioned region definitions to arrive at a more accurate and objective disease assessment for PM. We propose a deep learning-based approach to segment the abdomen into the 13~regions defined by the rPCI scoring system, thereby taking a first step toward automating the rPCI. The hypothesis is that automatic region division reduces interobserver variability between physicians when assessing the extent of PM, while also reducing the time required for evaluation and structuring the assessment process. The contributions of this work are threefold: (i)~the first deep learning approach for rPCI region segmentation, comparing a convolutional and a transformer-based architecture; (ii)~an anatomically constrained pipeline that encodes the geometric rPCI boundary definitions in a deterministic post-processing step; and (iii)~a comparison against interobserver agreement that places the achieved performance in clinical context.

\section{Methods}\label{sec:methods}
\subsection{Data Collection and Preprocessing}
This study was conducted using retrospectively collected imaging and clinical data, which were approved by the institutional review board of the Catharina Hospital in Eindhoven, the Netherlands. Imaging data were acquired from patients diagnosed with gastric, ovarian and colorectal cancer. Patients selected for inclusion underwent a CT scan, followed by DLS within six weeks of the CT scan to determine the sPCI. Moreover, patients did not have any concurrent treatment in between. The cohort included patients with sPCIs spanning the full range [0, 39]. A total of 62~CT scans were collected, all contrast-enhanced in the portal venous phase, with axial slice thicknesses ranging from thin (0.5--1.0~mm) to thick (3.0--5.0~mm).

\begin{figure}
   \centering
   \includegraphics[width=\textwidth]{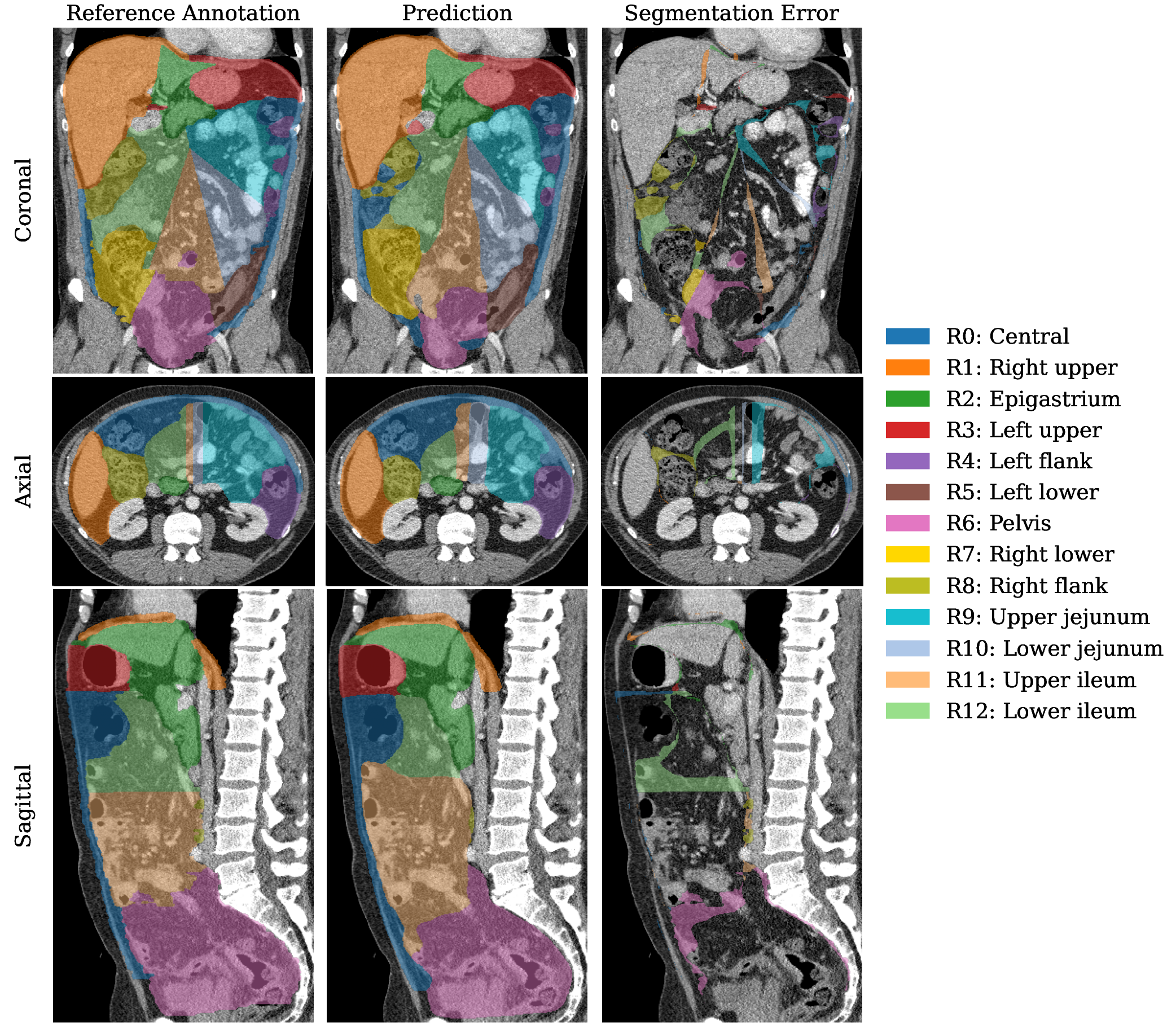}
   \caption{Visualization of CT scan slices from a patient with PM in multiple anatomical planes. The image is overlaid with the reference annotation~(left), the nnU-Net predicted mask~(middle) and the segmentation error~(right), showing false negative voxels that were present in the reference annotation but missed by nnU-Net.}
   \label{fig:XZ_gt_val_comparison}
\end{figure}
Ground-truth labels were produced by three clinical researchers, with every scan annotated once by a single annotator. The 62~scans were distributed according to annotator availability, resulting in 28, 17 and 17~scans per annotator. The annotations followed the rPCI region definitions as proposed by Tops-Welten~\emph{et~al.}~\cite{Tops-Welten2025DefiningStudy}. Each rPCI region was segmented as a 3D volume such that every voxel within the peritoneal cavity was either assigned to one region or the background. An example of a resulting reference annotation is shown in Fig.~\ref{fig:XZ_gt_val_comparison} (left column). Each completed annotation was subsequently reviewed by one of the other two annotators, and cases on which the two did not agree were referred to an expert radiologist, whose judgment was decisive. Interobserver agreement was therefore quantified on a separate set of scans that were each annotated independently by three observers, as described in Section~\ref{sec:interobserver}. The rPCI regions, numbered 0~to~12, are: central~(0), right upper~(1), epigastrium~(2), left upper~(3), left flank~(4), left lower~(5), pelvis~(6), right lower~(7), right flank~(8), upper jejunum~(9), lower jejunum~(10), upper ileum~(11), and lower ileum~(12)~\cite{Jacquet1996}.

During preprocessing, all CT scans were cropped to the segmentation mask boundaries while maintaining a margin of 15~mm in each dimension. This procedure excludes most of the thoracic region and background voxels, ensuring that Swin UNETR and nnU-Net were trained on the relevant abdominal region. Finally, to compensate for under-segmentation, the segmentation masks were dilated using a distance-based method with an expansion radius of 2~mm. In this case, a slight over-segmentation was preferred to ensure full anatomical coverage of each rPCI region.

\subsection{Model Selection}
nnU-Net and Swin UNETR were selected to represent two complementary and widely adopted paradigms for 3D medical image segmentation. nnU-Net is a self-configuring framework that reliably reaches strong performance on diverse biomedical datasets with minimal manual tuning by automatically adapting preprocessing, architecture, and training settings to a given dataset~\cite{Isensee2018NnU-Net:Segmentation}. The model provides a robust CNN baseline and an established reference built on the U-Net encoder–decoder with skip connections~\cite{ronneberger2015unetconvolutionalnetworksbiomedical}. In contrast, Swin UNETR employs a Swin Transformer encoder to model long-range dependencies through self-attention, which is particularly relevant for abdominal anatomy where structures span large spatial extents in 3D~\cite{Liu2021SwinWindows, Hatamizadeh2022SwinImages}. Evaluating both a strong CNN (nnU-Net) and a transformer-based approach (Swin UNETR) allows us to (i) assess whether long-range modeling benefits rPCI region delineation, (ii) compare performance stability across anatomically variable regions, and (iii) provide a reproducible baseline covering the two most commonly used families of segmentation architectures.

\subsection{Model Training}
nnU-Net and Swin UNETR were trained for 3D image segmentation in a supervised fashion using the annotated rPCI regions as ground truth labels. Training was performed on four NVIDIA A100 GPUs~(NVIDIA Corp., CA, USA), each with 40~GB of memory.
nnU-Net automatically optimizes its hyperparameters during preprocessing, resulting in a batch size of 2, patch size of $[96, 224, 112]$ voxels and a voxel spacing of $[1.73, 1.09, 1.73]$~mm for thin-slice acquisitions~(0.5--1.0~mm). Furthermore, the default nnU-Net pipeline was used, including CT normalization, network topology, SGD with Nesterov momentum (0.99), linear learning-rate decay, and its standard 3D data augmentations (random rotations and scaling, Gaussian noise and blur, brightness, contrast, and gamma augmentations, simulated low resolution, and mirroring along all spatial axes).

For Swin UNETR, we used a batch size of~1, a patch size of $[96, 96, 96]$, the same voxel spacing as nnU-Net, a learning rate of $1\times10^{-4}$, weight decay of $1\times10^{-5}$, an AdamW optimizer, and a Dice--Cross-Entropy loss. Preprocessing steps included intensity scaling, foreground cropping, reorientation, and resampling. Training augmentations consisted of balanced positive/negative patch sampling, random flips, 3D rotations ($\pm 15^\circ$), and random intensity shifts. Both models were evaluated using five-fold cross-validation with identical training and validation splits.

In addition to the baseline nnU-Net, an identically configured nnU-Net was trained on a merged label set consisting of three super-regions. The following regions were merged during training: left flank and left lower~(4$\cup$5), right flank and right lower~(7$\cup$8), and the small bowel regions~(9$\cup$10$\cup$11$\cup$12). Training on merged labels removes the need to learn boundaries that are less pronounced on CT; these super-regions are subsequently split back into the 13~rPCI regions by the anatomically constrained post-processing described below.

\subsection{Anatomically Constrained Post-processing}
Several rPCI boundaries are geometrically defined and more subtle on CT: the flank/lower boundaries~(4/5 and 7/8) and those separating the four small-bowel regions~(9--12). We address this with \emph{anatomically constrained post-processing}: the merged super-regions predicted by nnU-Net described above are split into the individual rPCI regions, using anatomical landmarks obtained from TotalSegmentator~\cite{Wasserthal2023TotalSegmentator}. Each left and right flank/lower super-region is divided by a transverse plane at the superior extent of the corresponding left or right hip bone. The small-bowel super-region is divided by anteroposterior planes radiating from the duodenum--small-bowel transition. Thereby, we approximate the mesenteric root at the ligament of Treitz, which is used as an anatomical landmark for constructing regions~9--12~\cite{Tops-Welten2025DefiningStudy}. From this point, four equal volume wedges are drawn. This post-processing is applied identically across all folds, and the resulting rPCI masks are evaluated with the same five-fold cross-validation and metrics as the baseline nnU-Net.

\subsection{Evaluation Metrics}
The Dice Similarity Coefficient (Dice) was used as the primary overlap metric and the 95th percentile Hausdorff distance (HD95, in mm) and Average~Surface~Distance (ASD, in mm) as boundary-accuracy metrics. Metrics were computed per patient and per region and summarized as mean~$\pm$~standard deviation across patients. Additionally, the dispersion was summarized using interquartile ranges and outlier counts. For clinical interpretability, overall (across-region) aggregates were also reported. In addition to model performance, interobserver variability was computed using the same metrics to contextualize the agreement between human annotators.

\subsection{Interobserver Reference}\label{sec:interobserver}
To quantify human variability, we computed interobserver agreement on a separate set of ten CT scans, organized as two cohorts of five, each independently annotated by three observers. Four clinical researchers contributed to this set: two annotated all ten scans, whereas the other two each annotated one cohort of five. In contrast to the training cohort, these scans were deliberately annotated in triplicate and the resulting masks were never reconciled into a single reference annotation. This subset was not used for training nnU-Net or Swin UNETR. Independent annotation of each scan by three observers resulted in three segmentation masks per scan, and therefore 30~segmentation~masks in total (10~scans $\times$ 3~observers). For each observer, we then computed Dice, HD95, and ASD between that observer's mask and the union of the remaining two observers' masks. Region-wise mean~$\pm$~std values were obtained by averaging these pairwise scores across observers and patients. Model agreement was quantified analogously: the predictions of the proposed anatomically constrained pipeline on these ten scans were compared against each of the three observer annotations and averaged, so that model-versus-human and human-versus-human agreement are measured on identical data. The interobserver metrics serve as a clinical reference for interpreting segmentation performance, noting that they are computed on a different dataset than the five-fold cross-validation results.

\section{Results and Discussion}\label{sec:results}
\begin{figure}
   \centering
   \includegraphics[width=0.95\textwidth]{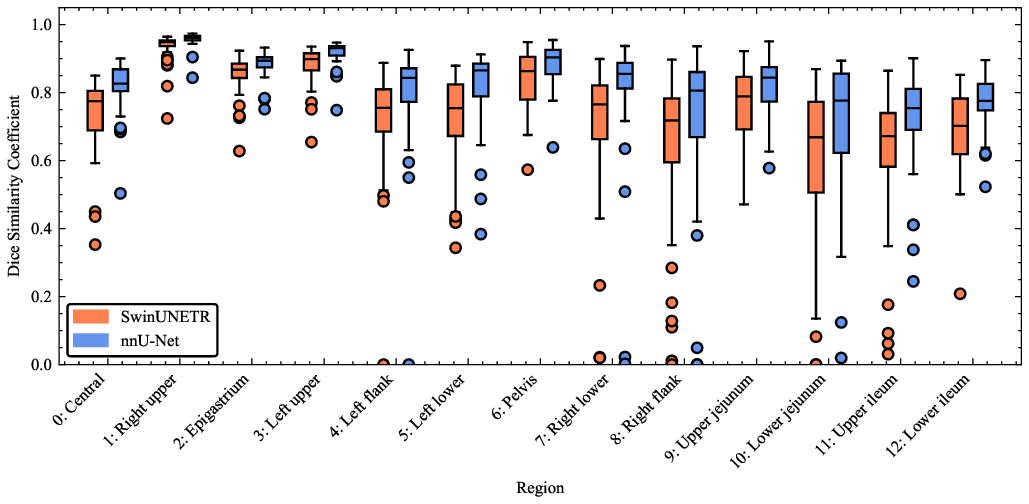}
   \caption{Per-region Dice distributions shown as boxplots for Swin UNETR (orange) and nnU-Net (blue). Outliers are indicated by equally colored dots. Both models exhibit a similar mean Dice across regions although nnU-Net shows much less variance across samples. Higher performance is observed for regions~0--7 and~9 and lower performance is observed for the remaining regions~8 and~10--12.}
   \label{fig:dice_comparison}
\end{figure}

Fig.~\ref{fig:dice_comparison} presents the Dice distributions for nnU-Net and Swin UNETR for each rPCI region. Averaged across all 13~regions, nnU-Net achieved a mean Dice of 0.81 compared to 0.76 for Swin UNETR, corresponding to a 6.6~\% relative improvement in overlap performance. In addition, nnU-Net has a higher median Dice value for every region, suggesting better typical regional segmentation performance across patients. The boxplots for regions 8 and 10 exhibit wide interquartile ranges and long whiskers, indicating high variability in performance across patients for both models. Furthermore, extreme outliers are observed for regions~7, 8, 10 and 11 in both models. Swin UNETR shows a larger spread compared to nnU-Net, indicating that it is more sensitive to patient-specific variability. Overall, nnU-Net shows more stable performance across the cohort of patients, with the exception of some outliers. Given its higher accuracy and stability, nnU-Net was used as the base model for the proposed anatomically constrained pipeline.

Fig.~\ref{fig:XZ_gt_val_comparison} illustrates a CT scan of the abdominal region of a patient in three orthogonal views, together with the reference annotation and predicted segmentation masks for the rPCI regions. The left column shows the reference annotations, the middle column presents the nnU-Net predictions, and the right column visualizes the segmentation error. Specifically, the error masks depict false negative voxels, defined as rPCI voxels present in the reference annotation but not predicted by nnU-Net. For regions~6, 8, 9 and 12, relatively large prediction errors are visible. Segmentation errors frequently occur near regional boundaries. Several factors may contribute to these errors. First, there is natural anatomical variability between patients. Second, the dataset includes patients in which organs have been partially or entirely surgically removed, as well as cases with large tumors that substantially alter normal anatomical relationships by displacing adjacent organs. In a relatively small dataset, these outliers can influence model optimization, thereby affecting model robustness and segmentation consistency across patients.

\begin{table}
   \centering
   \caption{Per-region comparison between the baseline nnU-Net (B) and the anatomically constrained pipeline (P), evaluated on the identical 62~CT scans using five-fold cross-validation. All values are mean~$\pm$~std across patients. Voxel\,\% denotes the percentage of foreground voxels per region. Dice (unitless), HD95 and ASD (mm). Region indices correspond to rPCI regions 0--12.}
   \label{tab:region_metrics_models}
   \footnotesize
   \setlength{\tabcolsep}{4pt}
   \begin{tabular}{lc cc cc cc}
      \toprule
       & & \multicolumn{2}{c}{Dice} & \multicolumn{2}{c}{HD95 (mm)} & \multicolumn{2}{c}{ASD (mm)} \\
      \cmidrule(lr){3-4}\cmidrule(lr){5-6}\cmidrule(lr){7-8}
      Region & Vox.\% & B & P & B & P & B & P \\
      \midrule
      0  & 14.85 & 0.82 $\pm$ 0.07 & 0.80 $\pm$ 0.09 &  9.2 $\pm$  5.5 & 13.9 $\pm$  7.8 & 2.4 $\pm$ 1.2 & 3.3 $\pm$ 1.6 \\
      1  & 19.45 & 0.96 $\pm$ 0.02 & 0.96 $\pm$ 0.02 &  6.4 $\pm$  4.1 &  5.9 $\pm$  3.7 & 1.6 $\pm$ 0.8 & 1.5 $\pm$ 0.7 \\
      2  &  6.14 & 0.89 $\pm$ 0.04 & 0.89 $\pm$ 0.03 &  8.8 $\pm$  6.2 &  8.1 $\pm$  5.8 & 2.5 $\pm$ 1.4 & 2.3 $\pm$ 1.3 \\
      3  & 12.93 & 0.92 $\pm$ 0.04 & 0.92 $\pm$ 0.05 &  8.4 $\pm$  6.6 &  7.6 $\pm$  5.5 & 2.1 $\pm$ 1.2 & 2.0 $\pm$ 1.2 \\
      4  &  3.15 & 0.81 $\pm$ 0.13 & 0.80 $\pm$ 0.14 & 13.9 $\pm$  8.7 & 13.4 $\pm$  8.9 & 3.5 $\pm$ 2.2 & 3.4 $\pm$ 2.4 \\
      5  &  2.95 & 0.82 $\pm$ 0.11 & 0.81 $\pm$ 0.13 & 12.7 $\pm$ 13.0 & 11.3 $\pm$ 10.9 & 3.4 $\pm$ 3.4 & 3.6 $\pm$ 4.4 \\
      6  & 12.24 & 0.89 $\pm$ 0.06 & 0.88 $\pm$ 0.06 & 12.0 $\pm$  7.2 & 12.8 $\pm$  8.1 & 3.4 $\pm$ 1.8 & 3.5 $\pm$ 1.9 \\
      7  &  3.60 & 0.80 $\pm$ 0.17 & 0.82 $\pm$ 0.17 & 14.1 $\pm$ 14.7 & 13.2 $\pm$ 14.0 & 5.3 $\pm$ 10.3 & 4.8 $\pm$ 9.0 \\
      8  &  3.59 & 0.72 $\pm$ 0.21 & 0.74 $\pm$ 0.20 & 17.3 $\pm$ 12.4 & 17.2 $\pm$ 16.5 & 5.8 $\pm$ 6.9 & 4.7 $\pm$ 5.3 \\
      9  &  5.35 & 0.81 $\pm$ 0.09 & 0.86 $\pm$ 0.07 & 16.5 $\pm$  9.4 & 11.7 $\pm$  5.8 & 4.7 $\pm$ 3.5 & 3.1 $\pm$ 1.4 \\
      10 &  5.21 & 0.69 $\pm$ 0.21 & 0.80 $\pm$ 0.15 & 20.7 $\pm$ 12.0 & 12.5 $\pm$  6.5 & 6.9 $\pm$ 4.7 & 4.0 $\pm$ 2.7 \\
      11 &  5.24 & 0.68 $\pm$ 0.18 & 0.81 $\pm$ 0.09 & 20.4 $\pm$ 12.3 & 12.3 $\pm$  6.4 & 6.7 $\pm$ 4.5 & 3.8 $\pm$ 2.1 \\
      12 &  5.31 & 0.77 $\pm$ 0.09 & 0.82 $\pm$ 0.08 & 17.3 $\pm$  8.6 & 13.6 $\pm$  7.5 & 5.0 $\pm$ 2.7 & 3.7 $\pm$ 1.7 \\
      \midrule
      Overall & - & 0.81 $\pm$ 0.15 & \textbf{0.84 $\pm$ 0.13} & 13.7 $\pm$ 10.8 & \textbf{11.8 $\pm$ 9.4} & 4.1 $\pm$ 4.6 & \textbf{3.4 $\pm$ 3.6} \\
      \bottomrule
      \end{tabular}
\end{table}

\begin{table}
   \centering
   \caption{Interobserver agreement and model performance comparison for rPCI region segmentation. Human metrics (Dice$_H$, HD95$_H$, ASD$_H$) show agreement between observers on ten CT scans with triple annotations (mean~$\pm$~std). Model metrics (Dice$_M$, HD95$_M$, ASD$_M$) show the proposed anatomically constrained pipeline against the observer annotations (mean~$\pm$~std). Dice (unitless), HD95 and ASD (mm).}
   \label{tab:interobserver_metrics}
   \begin{tabular}{lcccccc}
      \toprule
      Region & Dice$_H$ & Dice$_M$ & HD95$_H$ & HD95$_M$ & ASD$_H$ & ASD$_M$ \\
      \midrule
      0  & 0.85 $\pm$ 0.03 & 0.80 $\pm$ 0.06 &  6.6 $\pm$ 1.3 &  9.8 $\pm$ 4.5 & 1.9 $\pm$ 0.4 & 2.6 $\pm$ 0.8 \\
      1  & 0.96 $\pm$ 0.01 & 0.95 $\pm$ 0.01 &  5.3 $\pm$ 1.5 &  5.0 $\pm$ 1.9 & 1.3 $\pm$ 0.4 & 1.4 $\pm$ 0.5 \\
      2  & 0.91 $\pm$ 0.02 & 0.90 $\pm$ 0.02 &  6.6 $\pm$ 2.2 &  6.8 $\pm$ 2.1 & 1.6 $\pm$ 0.5 & 1.8 $\pm$ 0.4 \\
      3  & 0.94 $\pm$ 0.02 & 0.92 $\pm$ 0.01 &  5.0 $\pm$ 1.2 &  5.7 $\pm$ 1.4 & 1.3 $\pm$ 0.2 & 5.0 $\pm$ 3.2 \\
      4  & 0.86 $\pm$ 0.07 & 0.81 $\pm$ 0.09 &  9.9 $\pm$ 6.1 & 18.2 $\pm$ 21.3 & 2.3 $\pm$ 1.0 & 4.2 $\pm$ 4.4 \\
      5  & 0.86 $\pm$ 0.05 & 0.78 $\pm$ 0.15 &  5.4 $\pm$ 3.1 & 11.1 $\pm$ 12.4 & 1.6 $\pm$ 0.5 & 2.7 $\pm$ 2.1 \\
      6  & 0.90 $\pm$ 0.03 & 0.90 $\pm$ 0.03 &  9.2 $\pm$ 4.3 & 11.2 $\pm$ 9.1 & 2.4 $\pm$ 1.0 & 2.6 $\pm$ 1.2 \\
      7  & 0.84 $\pm$ 0.08 & 0.83 $\pm$ 0.08 &  8.4 $\pm$ 6.3 &  7.7 $\pm$ 4.8 & 2.2 $\pm$ 1.0 & 2.1 $\pm$ 1.0 \\
      8  & 0.83 $\pm$ 0.11 & 0.72 $\pm$ 0.15 & 12.1 $\pm$ 11.1 & 60.7 $\pm$ 64.4 & 2.8 $\pm$ 1.9 & 9.6 $\pm$ 8.0 \\
      9  & 0.84 $\pm$ 0.24 & 0.82 $\pm$ 0.07 & 16.9 $\pm$ 34.4 & 16.3 $\pm$ 6.8 & 7.5 $\pm$ 20.8 & 4.2 $\pm$ 1.6 \\
      10 & 0.83 $\pm$ 0.23 & 0.80 $\pm$ 0.08 & 11.2 $\pm$ 12.4 & 18.1 $\pm$ 10.7 & 3.5 $\pm$ 4.6 & 4.5 $\pm$ 1.8 \\
      11 & 0.83 $\pm$ 0.22 & 0.84 $\pm$ 0.09 & 10.4 $\pm$ 10.9 & 11.7 $\pm$ 6.8 & 3.3 $\pm$ 4.5 & 3.2 $\pm$ 1.6 \\
      12 & 0.83 $\pm$ 0.23 & 0.83 $\pm$ 0.11 & 17.0 $\pm$ 32.6 & 11.2 $\pm$ 7.6 & 7.2 $\pm$ 19.2 & 3.3 $\pm$ 2.2 \\
      \midrule
      Overall & 0.87 $\pm$ 0.07 & 0.84 $\pm$ 0.03 &  9.5 $\pm$ 6.5 & 14.9 $\pm$ 6.1 & 3.0 $\pm$ 3.7 & 3.6 $\pm$ 1.0 \\
      \bottomrule
   \end{tabular}
\end{table}

Table~\ref{tab:region_metrics_models} compares the baseline nnU-Net with the proposed anatomically constrained pipeline, evaluated on the identical 62~CT scans and folds, reported as Dice, HD95, and ASD. The baseline performs well in regions~1--3 and~6 but degrades in geometrically defined regions: right flank~(8) and the small bowel regions~(9--12). The proposed pipeline improves the overall Dice ($0.81 \pm 0.15 \to 0.84 \pm 0.13$) and reduces boundary error (HD95 $13.7 \to 11.8$~mm; ASD $4.1 \to 3.4$~mm), with gains concentrated in the small-bowel regions~(9--12) and right flank/lower regions~(7 and~8). Regions~4 and~5 show a slight decrease in Dice compared to the baseline nnU-Net. Apart from a minor regression in the central region~(0), which exhibits a decrease in Dice ($0.82 \to 0.80$), the non-post-processed regions remain essentially unchanged.

To contextualize these results, Table~\ref{tab:interobserver_metrics} reports interobserver agreement on a separate subset of ten CT scans, each independently annotated by three clinical researchers. Human (Dice$_H$, HD95$_H$, ASD$_H$) and model (Dice$_M$, HD95$_M$, ASD$_M$) agreement metrics were computed as described in Section~\ref{sec:interobserver}, the latter from predictions of the proposed anatomically constrained pipeline. The proposed pipeline achieves an overall Dice of 0.84 (model vs humans) compared to 0.87 for interobserver agreement. HD95 and ASD values are in the same order of magnitude (HD95 14.9~mm vs 9.5~mm; ASD 3.6~mm vs 3.0~mm). In most regions (e.g.,~0--3, 5--7 and~9--12) model performance approaches human agreement. For regions~9--12, the proposed model exhibits much less variation in performance across samples, whereas humans show high standard deviation with respect to each metric. The main exception is the right flank~(8), which shows a lower Dice and a high variance HD95. Inspection of interobserver scans revealed some outlier cases with deviating anatomy, which can negatively affect a rule-based post-processing.

The anatomically constrained pipeline yields the largest gains in the small-bowel regions~(9--12), substantially improving mean Dice ($0.74 \to 0.82$) and reducing ASD ($5.8 \to 3.7$). Residual errors are governed by landmark localization (e.g., the ligament of Treitz) rather than model training errors. Although the rPCI regions are defined according to established criteria, some variability in their delineation persists, as concluded in the consensus study by Tops-Welten~\emph{et~al.}~\cite{Tops-Welten2025DefiningStudy}. This variability indicates that small differences in boundary placement are expected and may not substantially affect downstream use of the PCI score. Consequently, the model might not be required to replicate perfect rPCI regions.

\subsection{Clinical Utility}
In current clinical practice, radiologists assess PM on imaging without standardized regional boundaries, contributing to interobserver variability and inconsistent reporting. Automatic rPCI region segmentation addresses this by providing a structured framework that (i)~guides radiologists to systematically inspect lesions within predefined anatomical regions, (ii)~reduces interobserver variability between clinicians, and (iii)~standardizes the evaluation of peritoneal lesions for radiology reports and surgical planning~\cite{Tops-Welten2025DefiningStudy, Fleurkens2026Impact}. Additionally, the structured overlay may accelerate reporting and support the training of less experienced readers. To realize these benefits in clinical practice, integration into the picture archiving and communication system is essential. On average, nnU-Net inference required approximately 7~seconds per scan on a single NVIDIA A100 GPU, suggesting computational feasibility for clinical integration.

Failure analysis indicates that most segmentation errors occur (a)~near regional borders and (b)~in patients with atypical anatomy (e.g., enlarged organs, removed organs or organ parts, large primary tumors), consistent with outliers seen in Fig.~\ref{fig:dice_comparison}. Importantly, qualitative corrections typically involved minor boundary adjustments rather than gross region reassignments. This suggests that a human-in-the-loop workflow, where the model proposes regions and radiologists refine boundaries as needed, can reduce the total delineation time while preserving clinical accuracy. Looking ahead, the segmented regions can support downstream automation of rPCI scoring by localizing lesions within each region and mapping lesion size to PCI subscores.

\subsection{Limitations and Generalizability}
This is an early, single-center study with a private dataset, which limits generalizability. CT scanner types, acquisition protocols, and reconstruction kernels vary across institutions and can affect model performance, particularly when interpatient anatomical variability is high. Furthermore, while all scans were acquired with intravenous contrast, variation in contrast phase and timing across the dataset may influence model robustness to non-standardized acquisition protocols. External validation on datasets originating from multiple centers is therefore required. Moreover, the interobserver quantification in this work was performed on a subset of ten CT scans that were each annotated independently by three observers, rather than on the full training cohort. Triple-annotating all 62~scans was infeasible at approximately five hours per scan per annotator. For statistical strength, the interobserver variability study should be executed on a larger subset of samples.

The collection of a PM dataset presents significant challenges that limit the availability of external validation data. PM nodules typically originate from ovarian, gastric, or colorectal cancers, making PM a relatively rare condition that requires specialized oncological expertise for diagnosis and treatment. Creating a dataset involves identifying patients with confirmed PM, obtaining appropriate imaging, and performing time-intensive manual segmentation of all rPCI regions by trained clinical experts.

Furthermore, the annotated rPCI regions were used as ground truth segmentation masks to train and validate both models. However, the rPCI regions are annotated using definitions from the consensus study by Tops-Welten~\emph{et~al.}~\cite{Tops-Welten2025DefiningStudy}, which may introduce some limitations due to inherent variability in anatomical boundaries of the consensus criteria. Notably, the lack of standardized regional definitions has contributed to significant interobserver variability, which may explain why only a minority of radiologists currently apply the PCI scoring system on imaging~\cite{Hafeez2023}.

\section{Conclusion}\label{sec:conclusion}
This study addressed whether deep learning models can automate the segmentation of CT images into rPCI regions defined through expert consensus, as a foundation for non-invasive PM assessment. Our results show that automated rPCI region segmentation is feasible: a baseline nnU-Net achieved an overall Dice of 0.81 and provided robust performance in most regions, with remaining challenges mainly in the geometrically defined small-bowel and flank regions. Motivated by the observation that these boundaries are definitional rather than appearance-based, we introduced an anatomically constrained pipeline that encodes the consensus geometry through deterministic, landmark-based post-processing. This improved the overall Dice to 0.84, approaching the interobserver agreement of 0.87, with some of the largest gains in regions that the baseline struggled with. The comparison with interobserver variability provided clinical context, showing that model performance approaches human-level agreement in several regions. These findings indicate that automated rPCI region segmentation can serve as a viable building block for imaging-based PCI assessment. By providing standardized, reproducible region maps, such models have the potential to reduce interobserver variability, accelerate radiological assessment, and support structured reporting and surgical planning. In future work, we plan to (i)~perform external validation across institutions, (ii)~quantify interobserver variability on a larger set of multi-annotated scans, (iii)~refine the anatomical landmark localization and robustness of post-processing for deviating anatomy, and (iv)~extend the models to handle magnetic resonance imaging.

\backmatter

\bmhead*{Acknowledgements}
The authors would like to thank the Catharina Cancer Institute from the Catharina Hospital in Eindhoven, the Netherlands, for providing medical imaging data, expert annotations and clinical guidance used in this study. This work was carried out in collaboration with GROW Research Institute for Oncology and Reproduction, Maastricht University, the Netherlands. Funding was granted by the Hanarth fund, which supports artificial intelligence research in the field of oncology. The authors also thank SURF (www.surf.nl) for their support in using the National Supercomputer Snellius.

\section*{Declarations}

\textbf{Funding:} This work was supported by the Hanarth fund, which supports artificial intelligence research in the field of oncology (\url{https://www.hanarthfonds.nl}).

\textbf{Conflict of interest/Competing interests:} The authors declare no competing interests.

\textbf{Ethics approval and consent to participate:} This study was conducted using retrospectively collected imaging and clinical data approved by the Catharina Cancer Institute from the Catharina Hospital in Eindhoven, the Netherlands. All methods were carried out in accordance with relevant guidelines and regulations.

\textbf{Consent for publication:} Not applicable.

\textbf{Data availability:} The datasets used during the current study are not publicly available.

\textbf{Materials availability:} Not applicable.

\textbf{Code availability:} Source code is available at \url{https://github.com/PieterGort/rpci-region-segmentation}. Trained nnU-Net model weights for the baseline 13-class segmentation and for the anatomically constrained pipeline (merged-label training) are provided as releases at \url{https://github.com/PieterGort/rpci-region-segmentation/releases}.

\textbf{Author contribution:} P.C.G. conceived the study, developed the methodology, performed the experiments, and wrote the manuscript. L.J.S.E., A.F.v.H, L.D.K., M.W.T.-W. provided clinical expertise and assisted with data annotation. Specifically, L.J.S.E., M.W.T.-W. and A.F.v.H. annotated the CT scans of the training cohort and reviewed each other's annotations, and together with L.D.K. produced the independent annotations used for the interobserver analysis. C.H.B.C. contributed to the technical implementation. J.N., M.J.L, M.D.P.L and I.H.J.T.H provided clinical oversight and validation, with J.N. resolving annotation disagreements. F.v.d.S. supervised the project and provided guidance throughout. All authors reviewed and approved the final manuscript.

\bibliography{references}

\end{document}